%% file: convexity_project.tex
\documentclass[12pt]{amsart}

\usepackage{amsmath,thmtools}
\usepackage{amsthm,amsfonts,amssymb}
\usepackage[usenames]{xcolor}
\usepackage{enumerate}
\usepackage[ruled,noline,noend]{algorithm2e}
\usepackage{fullpage}
\usepackage{hyperref}
\usepackage{color,soul}
\usepackage{mathtools}
\usepackage[makeroom]{cancel}
\usepackage{color,soul}
\usepackage{cite}
\usepackage{graphicx}

\theoremstyle{definition}

\theoremstyle{plain}

\usepackage[export]{adjustbox}

\usepackage{multicol}
\title{A Summary Of The Kernel Matrix, And How To Learn It Effectively Using Semidefinite Programming}
\author{Amir-Hossein Karimi} % \\ 20391321}

\newcommand{\zvec}[1]{\boldsymbol{#1}}
\newcommand{\zset}[1]{\big\{ #1 \big\}}

\newcommand{\zavspace}{\vspace{3mm}}
\newcommand{\zbvspace}{\vspace{5mm}}
\newcommand{\zcvspace}{\vspace{7mm}}

\begin{document}
\maketitle

\vspace{-10mm}

\input{abstract.tex}       % \newpage
\input{introduction.tex}       % \newpage
\vspace{-5mm}
\input{notation.tex}           % \newpage
\input{background_kernels.tex} \newpage
\input{background_sdp.tex}     % \newpage
\input{main_derivations.tex}   % \newpage
\input{conclusion.tex}         \newpage

\bibliographystyle{IEEEtran}
% \bibliographystyle{apalike}
% \bibliography{refs.bib}
\bibliography{refs.bib}

\newpage

\input{appendix_a.tex}         \newpage
\input{appendix_b.tex}         \newpage
\input{appendix_c.tex}         \newpage

% \end{multicols}

\end{document}

%% file: abstract.tex
\section{Abstract}

Kernel-based learning algorithms are widely used in machine learning for problems that make use of the similarity between object pairs. Such algorithms first embed all data points into an alternative space, where the inner product between object pairs specifies their distance in the embedding space. Applying kernel methods to partially labeled datasets is a classical challenge in this regard, requiring that the distances between unlabeled pairs must somehow be learnt using the labeled data. In this independent study, I will summarize the work of G. Lanckriet \textit{et al.}'s work \cite{lanckriet2004learning} on ``Learning the Kernel Matrix with Semidefinite Programming'' used in support vector machines (SVM) algorithms for the \textit{transduction} problem. Throughout the report, I have provide alternative explanations / derivations / analysis related to this work which is designed to ease the understanding of the original article.

%% file: introduction.tex
\section{Introduction}

Kernel methods \cite{christianini2000support, shawe2004kernel, scholkopf2002learning} are widely used in machine learning as they extend the application of algorithms designed for linear feature spaces to implicit non-linear feature spaces where the projected data is more separable. To meet this objective, kernel methods employ a positive-definite kernel function \mbox{$\Phi: \mathcal{X} \rightarrow \mathcal{F}, \mathcal{X} \in \mathbb{R}^d, \mathcal{F} \in \mathbb{R}^D$} to map the inputs to a high-dimensional (potentially infinite-dimenisonal) feature space where non-linear relations in the data is better captured relative to linear models in the input space. In this resultant feature space, various classification methods can be applied depending on the task at hand, for example, kernel support vector machine (SVM) \cite{cortes1995support} and kernel ridge regression \cite{saunders1998ridge}.

Despite the improved data separability provided by kernel methods, operating in these high-dimensional spaces (say to train an SVM) is costly, and so we face the \textit{curse of dimensionality}. To avoid the cost of explicitly working in the high-dimensional feature spaces, the well-known \textit{kernel trick} \cite{aizerman1964theoretical} is employed. Say $\bf{x}, \bf{y} \in \mathcal{X}$ are original points \mbox{$\in \mathbb{R}^d$} and \mbox{$\Phi(\bf{x}), \Phi(\bf{y}) \in \mathcal{F}$} are the projected points \mbox{$\in \mathbb{R}^D$}. Now, because our classifier depends on the similarity between the projected points \mbox{\big( i.e., $\big \langle \Phi(\bf{x}), \Phi(\bf{y}) \big \rangle$ \big)}, and because computing this is costly, we instead consider a non-linear mapping \mbox{$\Phi: \mathcal{X} \rightarrow \mathcal{F}$}, such that for all  \mbox{$\bf{x}, \bf{y} \in \mathbb{R}^d$}, \mbox{$\big \langle \Phi(\bf{x}), \Phi(\bf{y}) \big \rangle_{\mathcal{F}} = \text{K}(\bf{x},\bf{y})$} for some kernel $\text{K}(\bf{x},\bf{y})$. From here, we can simply learn a classifier \mbox{$\bf{H}: \bf{x} \rightarrow \bf{w}^T \Phi(\bf{x})$} for some $\bf{w} \in \mathcal{F}$.

The information specifying the inner products between each pair of points in the embedding space is contained in the so-called \textit{kernel matrix}, which is symmetric (due to the commutative property of distance between two points) and positive semidefinite (positive definite if all points are linearly independent). This matrix essential describes the geometry of the embedding space. The importance of this lies in the fact that since kernel-based learning algorithms extract all information needed from inner products of training data points in $\mathcal{F}$, there is no need to learn a kernel function $\Phi$ over the entire sample space to specify the embedding of a finite training dataset. Instead, the finite-dimensional kernel matrix (also known as a \textit{Gram matrix}) that contains the inner products of training points in $\mathcal{F}$ is sufficient. There are, of course, statistical model selection problems to be addressed within this approach; in particular ``which projected space efficiently linearly separates the data?'' and consequently ``which kernel should we use?''. The answers to these questions determines the elements of the kernel matrix. While prior knowledge can be used to aid this kernel selection process (to some degree of success), G. Lanckriet \textit{et al.} propose a method to learning the kernel matrix from data using semidefinite programming (SDP) techniques.

In their report, the main focus is on the problem of \textit{transduction} - the problem of completing the labeling of a partially labeled dataset. Keep in mind that the unlabeled samples of the dataset for which the algorithm is to find a label, are specified a priori. Once again, this suggests that the algorithm need only to learn a set of entries in the Gram matrix, and not the kernel function itself. This finite-dimensional kernel matrix uses the available labels to learn a good embedding by minimizing an objective function over the space of positive semidefinite kernels. This embedding should exhibit certain properties that aim to entangle the learning of labeled and unlabeled data. The resulting kernel matrix can then be employed as part of any algorithm that uses kernel matrices. One example that the authors discuss in detail is the support vector machine (SVM), where they show this method yields a polynomial time algorithm in the number of test examples, whereas Vapnik's original method for transduction \cite{gammerman1998learning} scales exponentially in the number of test examples.

\subsection{Statement of Intent}

In this paper, I shall summarize a formal proof for G. Lanckriet \textit{et al.}'s work \cite{lanckriet2004learning} on ``Learning the Kernel Matrix with Semidefinite Programming''. In Section 4, I formally define the concepts of kernel-based learning algorithms. In Section 5, we shall review the basics of semidefinite programming along with laying out dual formulations, in Section 6 we look into how SDP is used in finding the optimal kernel matrix. In this section, we also see a specific application of this framework for automatically learning the 2-norm soft margin parameter $\tau = 1 / C$. Finally, in Section 7, I shall summarize the works presented findings. I have attempted to prove and or elaborate on some skipped over derivations in the appendices.

%% file: notation.tex
\section{Notation}
In the derivations below, vectors are represented using bold lower-case notation, e.g, \mbox{$\zvec{v} \in \mathbb{R}^n$}, and scalar values represented using lower-case italics, e.g., $v_1, v_2, ..., v_n$. Matrices are represented using upper-case italics, e.g., \mbox{$X \in \mathbb{R}^{m \times n}$}. For a square, symmetric matrix $X$, $X \succeq 0$ means that $X$ is positive semidefinite, while $X \succ 0$ means that $X$ is positive definite. For a vector $\zvec{v}$, the notations $\zvec{v} \ge 0$ and $\zvec{v} > 0$ are understood componentwise.

%% file: background_kernels.tex
\section{Kernel Methods}

Kernel-based learning algorithms \cite{christianini2000support, shawe2004kernel, scholkopf2002learning} work by taking the data from the input space $\mathcal{X}$ and embedding into a feature space $\mathcal{F}$, then searching for linear relations in the resultant space. Kernel-based learning algorithms perform this embedding implicitly, forgoing the need to explicitly specify the function \mbox{$\Phi: \mathcal{X} \rightarrow \mathcal{F}$} and instead by requiring that pair-wise distances between each pair of points is specified (as given by their inner product). A primary advantange of this approach is that computing inner products in a feature space $\mathcal{F}$ is often much easier than computing the coordinates of the points themselves.

\noindent \textbf{Definition 1: Kernel Function} \textit{A} kernel \textit{is a function $k$, such that $k(\zvec{x}_i, \zvec{x}_j) = \langle \Phi(\zvec{x}_i), \Phi(\zvec{x}_j) \rangle$ $\forall ~ \zvec{x}_i, \zvec{x}_j \in \mathcal{X}$, where $\Phi$ is a mapping from $\mathcal{X}$ to an (inner product) feature space $\mathcal{F}$.}

\noindent \textbf{Definition 2: Kernel Matrix} \textit{A} kernel matrix \textit{is a is a square matrix $K \in \mathbb{R}^{p \times p}$ such that $K_{ij} = k(\zvec{x}_i, \zvec{x}_j)$ for some $\zvec{x}_1, \cdots, \zvec{x}_n \in \mathcal{X}$ and some kernel function k. The kernel matrix is also known as the Gram marix.}

In this paper and in the context of transduction, we are given a finite dataset $\mathcal{X}$, and so the kernel matrix is finite and can specified in the following simple manner

\noindent \textbf{Proposition 1} \textit{Every positive semidefinite and symmtetric matrix is a kernel matrix. Conversely, every kernel matrix is symmetric and positive semidefinite.\footnotemark}

\footnotetext{
  \textit{Proof:} Considering the general case of a kernel matrix, let $K_{ij} = K(\zvec{x}_i, \zvec{x}_j) = \langle \phi(\zvec{x}_i), \Phi(\zvec{x}_j) \rangle ~ \forall ~ i, j \in [l] $.
  \begin{align*}
                                        % K_{ij}                &= K(\zvec{x}_i, \zvec{x}_j) = \langle \phi(\zvec{x}_i), \Phi(\zvec{x}_j) \rangle ~ \forall ~ i, j \in [l]                                       \\
    \overset{\forall \zvec{v}}\implies  \zvec{v}^T K \zvec{v} &= \sum_{i,j = 1}^{l} \zvec{v}_i \zvec{v}_j K_{ij} = \sum_{i,j = 1}^{l} \zvec{v}_i \zvec{v}_j \langle \phi(\zvec{x}_i), \Phi(\zvec{x}_j) \rangle \\
                                                              &= \Big\langle \sum_{i = 1}^{l} \zvec{v}_i \phi(\zvec{x}_i), \sum_{j = 1}^{l} \zvec{v}_j \Phi(\zvec{x}_j) \Big\rangle                            \\
                                                              &= \bigg|\bigg| \sum_{i = 1}^{l} \zvec{v}_i \phi(\zvec{x}_i) \bigg|\bigg|^2 \ge 0                                                                \\
  \end{align*}
  \indent \indent as required. \qed
}

Kernel-based algorithms such as Support vector machines seek out solutions that are affine functions in the feature space $f(\zvec{x}) = \langle \zvec{w}, \Phi(\zvec{x}) \rangle + b$ for some weight vector $\zvec{w} \in \mathcal{F}$. As long as the weight vector $\zvec{w}$ can be expressed as a linear combination of the inputs ($\zvec{w} = \Sigma_{i = 1}^n \alpha_i \Phi(\zvec{x_i})$), $f$ can be expressed as

\begin{align*}
  f(\zvec{x}) &= \langle \zvec{w},                                   \Phi(\zvec{x}) \rangle + b \\
              &= \langle \Sigma_{i = 1}^n \alpha_i \Phi(\zvec{x_i}), \Phi(\zvec{x}) \rangle + b \\
              &= \Sigma_{i = 1}^n \alpha_i \langle \Phi(\zvec{x_i}), \Phi(\zvec{x}) \rangle + b \\
              &= \Sigma_{i = 1}^n \alpha_i k(\zvec{x_i}, \zvec{x}) + b                          \\
\end{align*}

\begin{figure}[tp]
  \centering
  \begin{tabular}{cc}
    \includegraphics[height=5.25 cm, valign=m]{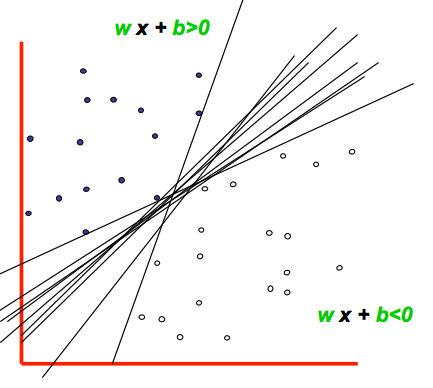} &
    \includegraphics[height=4.25 cm, valign=m]{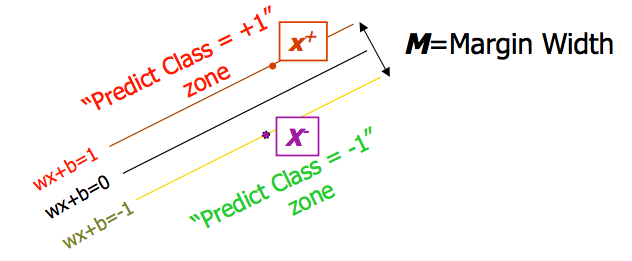}     \\
  \end{tabular}
  \vspace{-0.3 cm}
  \caption{(Left) Depicting various hyperplanes that can linearly separate the datapoints into two classes, only one of which separates is equidistant to the support vectors of both sides. (Right) SVMs are a class maximum-margin (i.e., $M = \gamma$ algorithms); the margin can be calculated as $\frac{\zvec{w}^T (\zvec{x}^+ - \zvec{x}^-)}{||\zvec{w}||_2} = \frac{2}{||\zvec{w}||_2}$}
  \label{fig:svm}
  % \vspace{-0.5 cm}
\end{figure}

For example in binary classification, a thresholded version of $f(\zvec{x})$, i.e. $sign \big( f(\zvec{x}) \big)$ is used to classify objects as belonging to one class or another. Note, in the examples and proofs that follow, we assume maximum-margin classifiers are learning the best separating hyperplanes for accurately separating objects from \textit{two} classes. As for multi-class margins, two prominent strategies commonly known as \textit{1-vs-all} or \textit{1-vs-1} are used, where each method is built upon kernel methods separating objects from 2 classes.

Furthermore, kernel-based algorithms such as SVMs are so-called maximum margin ($\gamma$) classifiers, trying to find the boundary that has the most distance between objects of different classes. In Figure \ref{fig:svm} we see various separating boundaries that cleanly divide objects from different classes into their corresponding buckets, but only the margin that has the maximum distance from objects of both class is chosen as the maximum-margin boundary, as it allows for more robustness in the face of new datapoints that may lie too close to the separating boundary. Geometrically, $\gamma$ corresponds to the distance between the convex hulls of the each of the two classes \cite{bennett2000duality} As is evident from Figure \ref{fig:svm}, the margin is inversly related to weights vector defining the separating boundary:

\begin{align*}
  \text{margin} = \gamma = \frac{2}{||\zvec{w}||_2}
\end{align*}

\noindent and so maximizing the margin is equivalent to minimizing this weight vector

\begin{align*}
  \underset{\zvec{w}, b}{\text{min}} \quad ||\zvec{w}||_2 = \underset{\zvec{w}, b}{\text{min}} \quad \langle \zvec{w}, \zvec{w} \rangle
\end{align*}

As discussed above, kernel methods attempt to find a function that is linear in the feature space $\mathcal{F}$ accurately separating instances of separate classes onto one or the other side of the function. In order for separating boundary to correctly find a boundary separating all samples, the optimization problem must optimize some criterion over all samples (i.e., each sample should somehow be included as a constraint). Here we look into several such criteria all of which are considered measures of separation of the labeled data. In addition to wanting to maximize the margin between objects of the two classes, kernel methods may allow a tolerance for misclassification. If no tolerance is allowed, the classifier is known as hard margin, whereas soft margin classifiers are those kernel methods that allow for misclassifications as incorporated in their optimization setup. Below we look into 3 criteria corresponding to different mesaures of separation / measures of performance.

\zbvspace
\noindent \textbf{Definition 3: Hard Margin} \textit{Given a set $S_l = \zset{(\zvec{x_1}, y_1), \cdots, (\zvec{x_n}, y_n)}$ with labeled examples, the hyperplane $(\zvec{w_*}, b_*)$ that solves the optimization problem below realizes the maximal margin classifier with geometric margin $\gamma = 1/ ||\zvec{w_*}||_2$ (assuming it exists). This margin is called the hard margin.}

\zcvspace

\noindent\begin{minipage}[t]{.5\linewidth}
  \textbf{Primal formulation:}
  \begin{equation*}
    \begin{aligned}
      \omega_H(K) = \underset{\zvec{w}, b}{\text{min}} & \quad \langle \zvec{w}, \zvec{w} \rangle                                   \\
      \textit{subject to}                              & \quad y_i \big( \langle \zvec{w}, \Phi(\zvec{x_i}) \rangle + b \big) \ge 1 \\
                                                       & \quad \forall ~ i \in [n]                                                  \\
    \end{aligned}
  \end{equation*}
\end{minipage}%
\begin{minipage}[t]{.5\linewidth}
  \textbf{Dual formulation\footnotemark:}
  \begin{equation*}
    \begin{aligned}
      \omega_H(K) = \underset{\zvec{\alpha}}{\text{max}} & \quad 2 \zvec{\alpha}^T \zvec{e} - \zvec{\alpha}^T G(K) \zvec{\alpha} \\
      \textit{subject to}                                & \quad 0 \le \zvec{\alpha}                                             \\
                                                         & \quad \zvec{\alpha}^T \zvec{y} = 0                                    \\
    \end{aligned}
  \end{equation*}
\end{minipage}
\footnotetext{Derivation in Appendix A.}

\zbvspace

\noindent where $G(K)$ is defined by $G_{ij}(K) = y_i y_j [K]_{ij} = y_i y_j k(\zvec{x_i}, \zvec{x_j})$, $\zvec{e}$ is the $n$-vector of ones, $\zvec{\alpha} \in \mathbb{R}^n$, and $\zvec{y} = [y_1, \cdots, y_n]^T$. The $n$ constraints in the primal follow straight from the fact that SVM would like to find a separable boundary that correctly (or with some margin of error) places each example on the proper side of the separable boundary; this means that each of the $n$ datapoints must be evaluated against the boundary that the optimzation problem finds.

We observe that the optimization problem finds a solution $\omega_H(K)$ to the hard margin problem if and only if the data is linearly separable in the transformed feature space $\mathcal{F}$ of the transformed datapoints, $\phi(\zvec{x_i})$. As seen in practice, this may not hold, and so we look for an alternative optimization problem that allows for a some error. For this, we define the 1-norm and 2-norm soft margin problems as follows.

\zbvspace

\noindent \textbf{Definition 4: 1-norm Soft Margin} \textit{Given a set $S_l = \zset{(\zvec{x_1}, y_1), \cdots, (\zvec{x_n}, y_n)}$ with labeled examples, the hyperplane $(\zvec{w_*}, b_*)$ that solves the optimization problem below realizes the maximal margin classifier with geometric margin $\gamma = 1/ ||\zvec{w_*}||_2$ (assuming it exists). This margin is called the 1-norm soft margin.}

\zcvspace

\noindent\begin{minipage}[t]{.5\linewidth}
  \textbf{Primal formulation:}
  \begin{equation*}
    \begin{aligned}
      \omega_{S1}(K) = \underset{\zvec{w}, b}{\text{min}} & \quad \langle \zvec{w}, \zvec{w} \rangle + C \Sigma_{i = 1}^n \eta_i                \\ % note I used \Sigma instead of \sum here because I wanted the minipages to be aligned (dual form did not have a sum)
      \textit{subject to}                                 & \quad y_i \big( \langle \zvec{w}, \Phi(\zvec{x_i}) \rangle + b \big) \ge 1 - \eta_i \\
                                                          & \quad \eta_i \ge 0                                                                  \\
                                                          & \quad \forall ~ i \in [n]                                                           \\
    \end{aligned}
  \end{equation*}
\end{minipage}%
\begin{minipage}[t]{.5\linewidth}
  \textbf{Dual formulation:}
  \begin{equation*}
    \begin{aligned}
      \omega_{S1}(K) = \underset{\zvec{\alpha}}{\text{max}} & \quad 2 \zvec{\alpha}^T \zvec{e} - \zvec{\alpha}^T G(K) \zvec{\alpha} \\
      \textit{subject to}                                   & \quad 0 \le \zvec{\alpha} \le C                                       \\
                                                            & \quad \zvec{\alpha}^T \zvec{y} = 0                                    \\
    \end{aligned}
  \end{equation*}
\end{minipage}

\zbvspace

\noindent \textbf{Definition 5: 2-norm Soft Margin} \textit{Given a set $S_l = \zset{(\zvec{x_1}, y_1), \cdots, (\zvec{x_n}, y_n)}$ with labeled examples, the hyperplane $(\zvec{w_*}, b_*)$ that solves the optimization problem below realizes the maximal margin classifier with geometric margin $\gamma = 1/ ||\zvec{w_*}||_2$ (assuming it exists). This margin is called the 2-norm soft margin.}

\zcvspace

\noindent\begin{minipage}[t]{.5\linewidth}
  \textbf{Primal formulation:}
  \begin{equation*}
    \begin{aligned}
      \omega_{S2}(K) = \underset{\zvec{w}, b}{\text{min}} & \quad \langle \zvec{w}, \zvec{w} \rangle + C \Sigma_{i = 1}^n \eta_i^2               \\ % note I used \Sigma instead of \sum here because I wanted the minipages to be aligned (dual form did not have a sum)
      \textit{subject to}                                 & \quad y_i \big( \langle \zvec{w}, \Phi(\zvec{x_i}) \rangle + b \big) \ge 1 - \eta_i  \\
                                                          & \quad \eta_i \ge 0                                                                   \\
                                                          & \quad \forall ~ i \in [n]                                                            \\
    \end{aligned}
  \end{equation*}
\end{minipage}%
\begin{minipage}[t]{.5\linewidth}
  \textbf{Dual formulation:}
  \begin{equation*}
    \begin{aligned}
      \omega_{S2}(K) = \underset{\zvec{\alpha}}{\text{max}} & \quad 2 \zvec{\alpha}^T \zvec{e} - \zvec{\alpha}^T \Big( G(K) +  1 / C I_n \Big) \zvec{\alpha} \\
      \textit{subject to}                                   & \quad 0 \le \zvec{\alpha}                                                                      \\
                                                            & \quad \zvec{\alpha}^T \zvec{y} = 0                                                             \\
    \end{aligned}
  \end{equation*}
\end{minipage}

\zbvspace

Intuitively, while the 2-norm loss penalizes outliers more than the 1-norm loss and so has slower convergance rate, it is both convex and smooth and so desirable to be used alongside a range of optimization problems.

\subsection{Generalized Performance Measure} \label{subsection_generalized_performance_measure}

The 3 definitions above have shown various formulations (primal \& dual) for finding the optimal separating margin $\gamma$. Noting the similarities in the 3 dual formulations above, we define the \textit{generalized performance measure} as

% \hl{(known as \textit{performance measure})}

\begin{equation*}
  \begin{aligned}
    \omega_{C,\tau}(K) = \underset{\zvec{\alpha}}{\text{max}} & \quad 2 \zvec{\alpha}^T \zvec{e} - \zvec{\alpha}^T \Big( G(K) +  \tau I_n \Big) \zvec{\alpha} \\
    \textit{subject to}                                       & \quad 0 \le \zvec{\alpha} \le C                                                               \\
                                                              & \quad \zvec{\alpha}^T \zvec{y} = 0                                                            \\
  \end{aligned}
\end{equation*}

\noindent with $\tau \ge 0$. Observe that each of the 3 formulations above are a special case of this general problem: $\omega_H(K) = \omega_{\infty, 0}$, $\omega_{S1}(K) = \omega_{C, 0}$, $\omega_{S2}(K) = \omega_{\infty, \tau = \frac1C}$. Given a kernel matrix $K$, this general optimization problem is able to find optimal parameters (i.e., $\zvec{w_*}, b_*$) that result in the maximum margin $\gamma$ between the samples, where pair-wise distances in the transformed feature space $\mathcal{F}$ are captured in $K$. As alluded to earlier, in the setting of \textit{transduction}, we are not given labels for all samples, and so we cannot possibly have an accurate $K$. We can easily see that to find the maximum margin separator in a transduction setting, we are first required to search over a set $\mathcal{K}$ of kernel matrices $K$, and find the maximum margin for each $K$. We can alternatively think of this as finding the minimum $\omega_{C,\tau}(K)$ over all possible $K$ matrices.

% Finding this maximum margin corresponds to minimizing $\omega_{C,\tau}(K)$ over the set of all possible kernel matrices $K$.

\vspace{-3mm}

\begin{equation*}
  \begin{aligned}
    \underset{K}{\text{min}} & \quad \omega_{C,\tau}(K)
  \end{aligned}
\end{equation*}

\zavspace

In the next sections, we shall describe the assumptions that assist in solving such a meta- / nested optimization problem. First, we will review semidefinite programs, a partitcular type of optimization problem with special structure for which exists theoretically and emperically efficient interior-point algorithms to solve.

%% file: background_sdp.tex
\section{Semidefinite Programming}

Semidefinite programming \cite{nesterov1994interior, vandenberghe1996semidefinite, boyd2004convex} deals with the optimization of convex functions over the convex cone of symmetric, positive semidefinite matrices

% \zcvspace

\begin{equation*}
  \begin{aligned}
    \mathcal{P} = \zset{X \in \mathbb{R}^{p \times p} \big| X = X^T, X \succeq 0}
  \end{aligned}
\end{equation*}

\zbvspace

\noindent or affine subsets thereof, characterized by affine or convex constraints on the optimization problem. As mentioned in the previous section, we aim to find an optimal kernel matrix $K$ for the task of transduction, and $\mathcal{P}$ (and convex subsets of $\mathcal{P}$) can be viewed as a search space for such possible kernel matrices. The key consideration in this paper is to specify a convex cost function that will enable us to find / learn the optimal kernel matrix using semidefinite programming.

\zbvspace

\noindent \textbf{Definition 4} \textit{A semidefinite program is a problem of the form}

% \zcvspace

\begin{equation*}
  \begin{aligned}
    \omega_{S2}(K) = \underset{\zvec{w}, b}{\text{min}} & \quad \zvec{c}^T \zvec{u}                                    \\
    \textit{subject to}                                 & \quad F^j(\zvec{u}) = F_0^j + u_1 F_1^j + \cdots + u_q F_q^j \\
                                                        & \quad A \zvec{u} = \zvec{b}                                  \\
                                                        & \quad \forall ~ j \in [L]                                    \\
  \end{aligned}
\end{equation*}

\zbvspace

\noindent \textit{where $\zvec{u} \in \mathbb{R}^q$ is the vector of decision variables, $\zvec{c} \in \mathbb{R}^q$ is the objective vector, and matrices $F_i^j = (F_i^j)^T \in \mathbb{R}^{p \times p}$ are given.}

An important property of each ``linear matrix inequality'' (LMI) $F^j(\zvec{u})$ is that the set of $\zvec{u}$ satisfying the LMI is a convex set. This property of each LMI constraint, along with the convexity of the affine constraint $A \zvec{u} = \zvec{b}$ leads to the SDPs being a convex optimization problem. This seemingly specialized form of SDPs is important as it arises in a host of applications in optimization problems (either directly in this form, or through casting problems as an SDP using the Schur's Complement Lemma\footnotemark), and because there exist theoretically and emperically efficient interior-point algorithms to solve SDPs \cite{vandenberghe1996semidefinite}.

\footnotetext{
  \noindent \textbf{Lemma 1 (Schur Complement Lemma)} \textit{Consider the partitioned symmetric matrix}

  \begin{equation*}
    \begin{aligned}
      X = X^T = \begin{pmatrix}
        A   & B \\
        B^T & C \\
      \end{pmatrix}
    \end{aligned}
  \end{equation*}

  % \zbvspace

  \noindent \textit{where $A, C$ are square and symmetric. If $\det(A) \not= 0$, we define the schur complement of $A$ in $X$ by the matrix $S = C - B^T A^{-1}B$. The Schur Complement Lemma states the following:}
  \begin{itemize}
    \item if $A \succ 0$, then \mbox{$X \succeq 0 \iff S \succeq 0$}
    \item \mbox{$X \succ 0 \iff A \succ 0 \textbf{ and } S \succ 0$}
  \end{itemize}
}

% \newpage

\subsection{SDP Duality, Optimality Conditions, \& Time Complexity} \label {subsection_sdp_duality}

The notion of duality in optimization problems can be extended to SDPs as well. Consider for example the following SDP with a single LMI constraint, an no affine equalities\footnotemark ~ can be expressed as

\footnotetext{In the general case, an SDP with multiple LMI constraints and affine equilities can be reduced to the aforementioned form by elimination of affine constraints, and merging all LMIs into 1 using block matrices.}

\zbvspace

\noindent\begin{minipage}[t]{.5\linewidth}
  \textbf{Primal formulation:}
  \begin{equation*}
    \begin{aligned}
      p^* = \underset{\zvec{u}}{\text{min}} & \quad \zvec{c}^T \zvec{u}                                      \\
      \textit{subject to}                   & \quad F(\zvec{u}) = F_0 + u_1 F_1 + \cdots + u_q F_q \preceq 0 \\
    \end{aligned}
  \end{equation*}
\end{minipage}%
\begin{minipage}[t]{.5\linewidth}
  \textbf{Dual formulation:}
  \begin{equation*}
    \begin{aligned}
      d^* &= \underset{Z \succeq 0}{\text{max}} \quad \underset{\zvec{u}}{\text{min}} \quad \mathcal{L}(\zvec{u}, Z)                   \\
          &= \underset{Z \succeq 0}{\text{max}} \quad \underset{\zvec{u}}{\text{min}} \quad \zvec{c}^T \zvec{u} + trace(Z F(\zvec{u})) \\
    \end{aligned}
  \end{equation*}
  % \textbf{Lagrangian formulation:}
  % \begin{equation*}
  %   \begin{aligned}
  %     p^* &= \underset{\zvec{u}}{\text{min}} \quad \underset{Z \succeq 0}{\text{max}} \quad \mathcal{L}(\zvec{u}, Z)                   \\
  %         &= \underset{\zvec{u}}{\text{min}} \quad \underset{Z \succeq 0}{\text{max}} \quad \zvec{c}^T \zvec{u} + trace(Z F(\zvec{u})) \\
  %         &= \begin{cases}
  %              % \item \zvec{c}^T \zvec{u} & \text{if } F(\zvec{u}) \preceq 0
  %              % \item +\infty             & \text{otherwise (i.e., infeasible region of primal).}
  %            \end{cases}
  %   \end{aligned}
  % \end{equation*}
\end{minipage}

\zbvspace

\noindent where $Z$ is a symmetric matrix, of the same size as $F(\zvec{u})$. Observe that the Lagrangian function $\mathcal{L}(\zvec{u}, Z) = \zvec{c}^T \zvec{u} + trace(Z F(\zvec{u}))$ does indeed fullfill the same objective as the primal SDP formulation as we can see that

\vspace{-3mm}

\begin{align*}
  \underset{Z \succeq 0}{\text{max}} = \begin{cases}
                                         \zvec{c}^T \zvec{u} & \text{if } F(\zvec{u}) \preceq 0                      \\
                                         +\infty             & \text{otherwise (i.e., infeasible region of primal).} \\
                                       \end{cases}
\end{align*}

\noindent where we have shown the above is a \textit{barrier} for the primal SDP. As in non-SDP duality, we always have weak duality and can exchange the $min$ and $max$ to get

\vspace{-3mm}

\begin{align*}
  d^* = \underset{Z \succeq 0}{\text{max}} \quad \underset{\zvec{u}}{\text{min}} \quad \mathcal{L}(\zvec{u}, Z) \quad \le \quad \underset{\zvec{u}}{\text{min}} \quad \underset{Z \succeq 0}{\text{max}} \quad \mathcal{L}(\zvec{u}, Z) = p^*
\end{align*}

\noindent We further have strong duality with a Slater-type condition stating that if the primal SDP is strictly feasible (i.e., $\exists ~ \zvec{u_0} ~ s.t. ~ F(\zvec{u_0}) \prec 0$) then $p^* = d^*$. Additionally, if the dual SDP\footnotemark ~ is also strictly feasible (i.e., $\exists ~ Z \succ 0 ~ s.t. ~ c_i = - trace(ZF_i) ~ \forall ~ i \in [q]$), then both primal and dual optimal values are attained by the same optimal pair $(\zvec{u^*}, Z^*)$.

\footnotetext{
  The dual is as follows:
  \begin{align*}
    d^* &= \underset{Z \succeq 0}{\text{max}} \quad \underset{\zvec{u}}{\text{min}} \quad \mathcal{L}(\zvec{u}, Z)                   \\
        &= \underset{Z \succeq 0}{\text{max}} \quad \underset{\zvec{u}}{\text{min}} \quad \zvec{c}^T \zvec{u} + trace(Z F(\zvec{u})) \\
        &= \underset{Z \succeq 0}{\text{max}} \quad \underset{\zvec{u}}{\text{min}} \quad \zvec{c}^T \zvec{u} + trace(Z F_0) + \sum_{i = 1}^q u_i trace(Z F_i) \\
        &= \underset{Z \succeq 0}{\text{max}} \begin{cases}
                                                trace(Z F_0) & \text{if } c_i = -trace(Z F_i) ~ \forall ~ i \in [q] \\
                                                -\infty      & \text{otherwise.}                                    \\
                                              \end{cases}
  \end{align*}

  \noindent which can be re-written as the following problem.

  \begin{equation*}
    \begin{aligned}
      d^* = \underset{Z}{\text{max}} & \quad trace(Z F_0)        \\
      \textit{subject to}            & \quad Z \succeq 0         \\
                                     & \quad c_i = -trace(Z F_i) \\
                                     & \quad \forall ~ i \in [q] \\
    \end{aligned}
  \end{equation*}

  \noindent which itself is an SDP. So the dual of a primal SDP is an SDP. We explain why this is important in the text.
}

We have seen that the primal and dual pair of an SDP are both SDPs, and assuming strict feasibility of both problems, solving the primal minimization problem and maximizing the dual problem are equivalent. This relationship is indeed what algorithms such as SeDuMi \cite{sturm1999using} exploit to simultaneously solve both the primal and dual, and then use the duality gap $p^* - d^*$ as a stopping criterion. Equipped with interior-point methods \cite{nesterov1994interior} for solving SDPs, these methods have a worst-case complexity of $O(q^2 p^{2.5})$ where $p$ refers to the dimensionality of $F(\zvec{u}) \in \mathbb{R}^{p \times p}$ and $q$ is the number of block matrices in the reduced LMI. In subsequent sections, we will explain how this bound may be improved assuming certain constraints on the SDP.

%% file: main_derivations.tex
\newcommand{\zweightedsigmaki}{\Sigma_{i = 1}^m \mu_i K_i}
\newcommand{\zweightedsigmakitr}{\Sigma_{i = 1}^m \mu_i K_{i, tr}}

\section{Algorithms for Learning Kernels} \label{section_algorithms_for_learning_kernels}

Working in a transduction setting, we are given data that is partially labeled; the training set $S_{n_{tr}} = \zset{(\zvec{x}_1, y_1), \cdots, (\zvec{x}_{n_{tr}}, y_{n_{tr}})}$ is labeled, and the remainder (i.e., test set) $T_{n_{t}} = \zset{\zvec{x}_{n_{tr} + 1}, \cdots, \zvec{x}_{n_{tr} + n_{t}}}$ are unlabeled, and the aim is to predict the labels for the test samples. In such a setting, the optimizing the kernel is equivalent to choosing a kernel matrix of form

\begin{equation*}
  \begin{aligned}
    K = \begin{pmatrix}
      K_{tr}     & K_{tr,t} \\
      K_{tr,t}^T & K_{t}    \\
    \end{pmatrix}
  \end{aligned}
\end{equation*}

\zavspace

\noindent where $K_{ij} = \langle \Phi(\zvec{x}_i), \Phi(\zvec{x}_j) \rangle, i,j \in [n_{tr} + n_{t}]$. By setting up the optimization problem to optimize over the ``training-data block'' $K_{tr}$, we want to learn the optimal ``mixed-data block'' $K_{tr,t}$ and ``test-data block'' $K_{t}$. What we mean to say is that training and test data blocks must somehow be entangled: tuning the entries in $K$ that correspond to training data (i.e., optimizing their embedding) should automatically tune the test-data entries in some way as well.  This can be achieved by constraining the search for optimal kernel to only a certain category of kernels to prevent overfitting the training data, and generalize well to unlabeled test data. For this, the authors consider 3 specific example search spaces $\mathcal{K}$:

\zcvspace

\noindent\begin{minipage}[t]{.35\linewidth}
  \textbf{Example 1:}
  \begin{equation*}
    \begin{aligned}
      K           &\in \mathcal{K}                               \\
      \mathcal{K} &\subseteq \mathcal{P} \text{ (convex subset)} \\
      \mathcal{P} &\text{: the PSD cone}                         \\
    \end{aligned}
  \end{equation*}
\end{minipage}%
\begin{minipage}[t]{.33\linewidth}
  \textbf{Example 2:}
  \begin{equation*}
    \begin{aligned}
      K        &= \zweightedsigmaki \\
      K        &\succeq 0           \\
      trace(K) &\le c               \\
    \end{aligned}
  \end{equation*}
\end{minipage}
\begin{minipage}[t]{.33\linewidth}
  \textbf{Example 3:}
  \begin{equation*}
    \begin{aligned}
      K        &= \zweightedsigmaki          \\
      \mu_i    &\ge 0 ~ \forall ~ i \in [m]  \\
      K        &\succeq 0                    \\
      trace(K) &\le c                        \\
    \end{aligned}
  \end{equation*}
\end{minipage}

\zbvspace

\noindent where in examples 2 \& 3, the set $\mathcal{K} = \zset{K_1, \cdots, K_m}$ are initial ``guesses" of the optimal kernel matrix $K = \zweightedsigmaki$. These initial guesses, i.e., $K_i$s, can be selected from the class of well-known kernel matrices, e.g., linear, Gaussian or polynomial kernels with different kernel parameter values (e.g., slack variables, Gaussian kernel width, polynomial kernel rank, etc.). Elements of $\mathcal{K}$ could be chosen as rank-one matrices $K_i = \zvec{v_i} \zvec{v_i}^T$ with $\zvec{v_i}$ some set of orthogonal vectors, or alternatively $K_i$s can be a diverse set of \textit{possibly good} Gram matrices using various heterogeneous data sources. In any case, the set of such $K_i$s, i.e., $\mathcal{K}$, lies on the intersection of a low-dimensional linear subspace with the positive semidefinite cone $\mathcal{P}$. Note, the constraints $\mathcal{K} \subseteq \mathcal{P}$ and $K \succeq 0$ (identical constraints) both arise from \mbox{Proposition 1} where kernel matrices are positive semidefinite by definition.

% \hl{Furthermore, bounding the trace allows for...}

% \hl{why does this entangle?!?!} \hl{all these sets $\mathcal{K}$ are convex?? is it important for an SDP? general optimization problem is convex in K... but does K itself have to be convex too?}.

Formulating the optimal kernel $K$ as a linear combination of $K_i$s has the interesting consequence that the optimization problem that tries to learn $K$ need only to find optimal mixing parameters $\mu_i ~ \forall ~ i \in [m]$, instead of fine-tuning the parameters of each of $K_i$ using cross-validation. We shall see that restricting the mixing parameters to be non-negative leads to significantly reduced computational complexity for the optimization problem.

% and is easier for studying the statistical properties of the kernel matrices.}

\subsection{General Optimization Problem} \label{subsection_general_optimization_problem}

Recall from Section \ref{subsection_generalized_performance_measure} where we derived \textit{generalized performance measure} (i.e., the general dual formulation) for the optimization problem when using any of the kernel methods (i.e., hard margin, 1-norm soft margin or 2-norm soft margin) for transduction. This had the form

\begin{equation*}
  \begin{aligned}
    \omega_{C,\tau}(K) = \underset{\zvec{\alpha}}{\text{max}} & \quad 2 \zvec{\alpha}^T \zvec{e} - \zvec{\alpha}^T \Big( G(K) +  \tau I_n \Big) \zvec{\alpha} \\
    \textit{subject to}                                       & \quad 0 \le \zvec{\alpha} \le C                                                               \\
                                                              & \quad \zvec{\alpha}^T \zvec{y} = 0                                                            \\
  \end{aligned}
\end{equation*}

An important property of the generalized performance measure above is its convexity in $K$. This can be asserted by considering first that the objective function $2 \zvec{\alpha}^T \zvec{e} - \zvec{\alpha}^T ( G(K) +  \tau I_n ) \zvec{\alpha}$ is an affine (and hence convex) function of $K$. Secondly, $\omega_{C,\tau}(K)$ is the pointwise maximum of such conex functions, and so is also convex. The remainder of the constraints $0 \le \zvec{\alpha} \le C$ and $\zvec{\alpha}^T \zvec{y} = 0$ are obviously convex. Hence the general optimization problem below that minimizes the performance measure over a family $\mathcal{K}$ of kernel matrices $K$ is convex.

Recall that in an earlier part of Section \ref{section_algorithms_for_learning_kernels} (this section), we described 3 examples for how to entangle the learning of the training-data and test-data blocks to learn labels for the unlabeled samples by imposing additional constraints on $\mathcal{K}$. These special forms of $\mathcal{K}$ allow us to cast and solve the general optimization problem efficiently as the following SDP\footnote{Derivation in Appendix B.}

\zavspace
% \zcvspace

\noindent\begin{minipage}[t]{.425\linewidth}
  \textbf{General optimization problem:}
  \begin{equation*}
    \begin{aligned}
      \underset{K}{\text{min}} & \quad \omega_{C,\tau}(K) \\
      \textit{subject to}      & \quad trace(K) = c       \\
    \end{aligned}
  \end{equation*}
\end{minipage}%
\begin{minipage}[t]{.6\linewidth}
  \textbf{Standard SDP Form:}
  \begin{equation*}
    \begin{aligned}
      \underset{K, \zvec{\nu}, \zvec{\delta}, \lambda, t}{\text{min}} & \quad t                                                                                                                      \\
      \textit{subject to}                                             & \quad trace(K) = c,                                                                                                          \\
                                                                      & \quad K \in \mathcal{K}                                                                                                      \\
                                                                      & \quad \begin{pmatrix}
                                                                                G(K_{tr}) + \tau I_{n_{tr}}                               & \zvec{e} + \zvec{\nu} - \zvec{\delta} + \lambda \zvec{y} \\
                                                                                (\zvec{e} + \zvec{\nu} - \zvec{\delta} + \lambda \zvec{y})^T & t - 2C \zvec{\delta}^T \zvec{e}                       \\
                                                                              \end{pmatrix} \succeq 0                                                                                                \\
                                                                      & \quad \zvec{\nu} \ge 0                                                                                                       \\
                                                                      & \quad \zvec{\delta} \ge 0                                                                                                    \\
    \end{aligned}
  \end{equation*}
\end{minipage}

% \zbvspace

\noindent where $\lambda \in \mathbb{R}$ and $\zvec{\nu}, \zvec{\delta} \in \mathbb{R}^{n_{tr}}$. We note that $\zvec{\nu} \ge 0 \iff diag(\zvec{\nu}) \succeq 0$ which means that contraint is an LMI and a proper SDP constraint; likewise for $\zvec{\nu} \ge 0$.

We now look into the 3 specific forms of $\mathcal{K}$ described above and compare performances and expected outcome.

\subsection{Example 1: $\mathcal{K} = \zset{K \succeq 0}$} \label{subsection_example_1}

\hfill \\

We notice that if $\mathcal{K} = \zset{K \succeq 0}$, the optimization problem is indeed an SDP in standard form, however, there are no constraints guaranteeing entanglement of learning training-data and test-data blocks. For example, $K_t = \mathcal{O}$ where $K_t$ is the test-data block and $\mathcal{O}$ denotes the zero matrix satisfies all constraints and is a viable solution to the minimization problem, yet it does not carry any meaning in terms of labels for unlabeled data. Below, we look into other kernel families $\mathcal{K}$ that can enforce the desired entangled learning.

\subsection{Example 2: $\mathcal{K} = span \zset{K_1, \cdots, K_m} \cap \zset{K \succeq 0}$} \label{subsection_example_2}

\hfill \\ % so new line after subsection

\noindent\begin{minipage}[t]{.4\linewidth}
  \textbf{Optimization problem:}
  \begin{equation*}
    \begin{aligned}
      \underset{K}{\text{min}} & \quad \omega_{C,\tau}(K_{tr}) \\
      \textit{subject to}      & \quad trace(K) = c            \\
                               & \quad K \succeq 0             \\
                               & \quad K = \zweightedsigmaki   \\
    \end{aligned}
  \end{equation*}
\end{minipage}%
\begin{minipage}[t]{.6\linewidth}
  \textbf{Standard SDP Form:}
  \begin{equation*}
    \begin{aligned}
      \underset{\zvec{\mu}, \zvec{\nu}, \zvec{\delta}, \lambda, t}{\text{min}} & \quad t                                                                                                                          \\
      \textit{subject to}                                                      & \quad trace(\zweightedsigmaki) = c,                                                                                              \\
                                                                               & \quad \zweightedsigmaki \succeq 0                                                                                                \\
                                                                               & \quad \begin{pmatrix}
                                                                                         G(\zweightedsigmakitr) + \tau I_{n_{tr}}                      & \zvec{e} + \zvec{\nu} - \zvec{\delta} + \lambda \zvec{y} \\
                                                                                         (\zvec{e} + \zvec{\nu} - \zvec{\delta} + \lambda \zvec{y})^T & t - 2C \zvec{\delta}^T \zvec{e}                           \\
                                                                                       \end{pmatrix} \succeq 0                                                                                                    \\
                                                                               & \quad \zvec{\nu} \ge 0                                                                                                           \\
                                                                               & \quad \zvec{\delta} \ge 0                                                                                                        \\
    \end{aligned}
  \end{equation*}
\end{minipage}

Solving this optimization problem in its standard SDP form can be done using general-purpose programs such as SeDuMi \cite{sturm1999using}. As discussed in Section \ref{subsection_sdp_duality}, the interior-point methods used to solve SDP problems are polynomial in time, however, have worst case complexity $O \big( (m + n_{tr})^2) (n^2 + n_{tr}^2) (n + n_{tr})^{0.5} \big)$ or roughly $O \big( (m + n_{tr})^2 n^{2.5} \big)$ with the assumptions in example 2. This itself is a major improvement over Vapnik's original method for transduction \cite{gammerman1998learning} that scales exponentially in the number of test examples.

\subsection{Example 3: $\mathcal{K} = span \zset{K_1, \cdots, K_m} \cap \zset{K \succeq 0}$ \& positive linear combinations} \label{subsection_example_3}

\zavspace
% \hfill \\

With the additional assumption on $\zweightedsigmaki$ that $\zvec{\mu} \ge 0$, we can derive the following\footnote{Derivation in Appendix C.}

\zavspace

\noindent\begin{minipage}[t]{.4\linewidth}
  \textbf{Optimization problem:}
  \begin{equation*}
    \begin{aligned}
      \underset{K}{\text{min}} & \quad \omega_{C,\tau}(K_{tr}) \\
      \textit{subject to}      & \quad trace(K) = c            \\
                               & \quad K \succeq 0             \\
                               & \quad K = \zweightedsigmaki   \\
                               & \quad \zvec{\mu} \ge 0        \\
    \end{aligned}
  \end{equation*}
\end{minipage}%
\begin{minipage}[t]{.6\linewidth}
  \textbf{Standard SDP Form:}
  \begin{equation*}
    \begin{aligned}
      \underset{\zvec{\alpha}, t}{\text{max}} & \quad 2 \zvec{\alpha}^T \zvec{e} - \tau \zvec{\alpha}^T \zvec{\alpha} - c t \\
      \textit{subject to}                     & \quad t \ge 1/r_i \zvec{\alpha}^T G( K_{i, tr} ) \zvec{\alpha}              \\
                                              & \quad \zvec{\alpha}^T \zvec{y} = 0                                          \\
                                              & \quad 0 \le \zvec{\alpha} \le C                                             \\
                                              & \quad \forall ~ i \in [m]                                                   \\
    \end{aligned}
  \end{equation*}
\end{minipage}

\noindent where $\zvec{r} \ in \mathbb{R}^m$ and $trace(K_i) = r_i$. This SDP form is more special. In fact, it is a quadratically-constrained quadratic probem (QCQP), itself a type of second-order cone program, which is a special type of SDP \cite{boyd2004convex}. QCQPs can be solve efficiently using either of the SeDuMi \cite{sturm1999using} Mosek \cite{andersen2000mosek} frameworks, which yield worst-case complexity $O \big( m n_{tr}^3 \big)$. This is a major improvement over the worst-case complexity of the previous example that didn't assume \textit{positive} linear combinations, and over the general SDP form where $K \succeq 0$. Additionally this constraint results in improved numerical stability - it prevents the algorithm from finding solutions that use large weights $\mu_i$ with opposite sign that cancel.

Observe that each of the 3 margin criteria (i.e., hard margin, 1-norm soft margin or 2-norm soft margin) is a special case of the generalized performance measure $\omega_{C,\tau}(K_{tr})$: $\omega_H(K) = \omega_{\infty, 0}$ (hard margin), $\omega_{S1}(K) = \omega_{C, 0}$ (1-norm soft margin), $\omega_{S2}(K) = \omega_{\infty, \tau = \frac1C}$ (2-norm soft margin). These can be used in conjunction with the 3 assumptions on the kernel family $\mathcal{K}$ to get appropriate SDP setups for different margins based on the problem at hand. Here we consider 1 example, that was an important open problem at the time G. Lanckriet \textit{et al.} were publishing their work.

\subsection{Learning the 2-norm Soft Margin Parameter}

This section shows how the 2-norm soft margin parameter $\tau = 1/C$ of SVMs can be learned using SDP or QCQP. For details, please refer to the work of the authors \cite{lanckriet2004learning} or those of De Bie \textit{et al.} \cite{de2003convex}. Recall the specific form of the 2-norm soft margin problem:

\noindent\begin{minipage}[t]{.5\linewidth}
  \textbf{Primal formulation:}
  \begin{equation*}
    \begin{aligned}
      \omega_{S2}(K) = \underset{\zvec{w}, b}{\text{min}} & \quad \langle \zvec{w}, \zvec{w} \rangle + C \Sigma_{i = 1}^n \eta_i^2               \\ % note I used \Sigma instead of \sum here because I wanted the minipages to be aligned (dual form did not have a sum)
      \textit{subject to}                                 & \quad y_i \big( \langle \zvec{w}, \Phi(\zvec{x_i}) \rangle + b \big) \ge 1 - \eta_i  \\
                                                          & \quad \eta_i \ge 0                                                                   \\
                                                          & \quad \forall ~ i \in [n]                                                            \\
    \end{aligned}
  \end{equation*}
\end{minipage}%
\begin{minipage}[t]{.5\linewidth}
  \textbf{Dual formulation:}
  \begin{equation*}
    \begin{aligned}
      \omega_{S2}(K) = \underset{\zvec{\alpha}}{\text{max}} & \quad 2 \zvec{\alpha}^T \zvec{e} - \zvec{\alpha}^T \Big( G(K) +  \tau I_n \Big) \zvec{\alpha} \\
      \textit{subject to}                                   & \quad 0 \le \zvec{\alpha}                                                                     \\
                                                            & \quad \zvec{\alpha}^T \zvec{y} = 0                                                            \\
    \end{aligned}
  \end{equation*}
\end{minipage}

\noindent where $\tau = 1 / C$. Remember $\tau = 0$ for both the hard margin and 1-norm soft-margin cases. Since in the dual formulation above, the identity matrix induced by the 2-norm appears in exactly the same wasy as other matrices in $G(K)$, we can treat them in the same way and jointly optimize over $K$ and $\tau$ to find the optimal dual formulation, i.e., minimize $\omega_{S2}(K_{tr})$. Now the joint optimization problem is as follows, and the dual can be restated as the following SDP (similar to derivations in Section \ref{subsection_general_optimization_problem}):

\noindent\begin{minipage}[t]{.425\linewidth}
  \textbf{General optimization problem:}
  \begin{equation*}
    \begin{aligned}
      \underset{K, \tau \ge 0}{\text{min}} & \quad \omega_{C,\tau}(K_{tr}, \tau) \\
      \textit{subject to}                  & \quad trace(K) = c                  \\
    \end{aligned}
  \end{equation*}
\end{minipage}%
\begin{minipage}[t]{.6\linewidth}
  \textbf{Standard SDP Form:}
  \begin{equation*}
    \begin{aligned}
      \underset{K, \tau, \zvec{\nu}, \lambda, t}{\text{min}} & \quad t                                                                                         \\
      \textit{subject to}                                    & \quad trace(K + \tau I_n) = c,                                                                  \\
                                                             & \quad K \in \mathcal{K}                                                                         \\
                                                             & \quad \begin{pmatrix}
                                                                       G(K_{tr}) + \tau I_{n_{tr}}                  & \zvec{e} + \zvec{\nu} + \lambda \zvec{y} \\
                                                                       (\zvec{e} + \zvec{\nu} + \lambda \zvec{y})^T & t                                        \\
                                                                     \end{pmatrix} \succeq 0                                                                   \\
                                                             & \quad \zvec{\nu} \ge 0                                                                          \\
                                                             & \quad \zvec{\delta} \ge 0                                                                       \\
    \end{aligned}
  \end{equation*}
\end{minipage}

\noindent Now we impose assumptions on the family $\mathcal{K}$ just as we did in example 2 (Section \ref{subsection_example_2}) and example 3 (Section \ref{subsection_example_3}):

\noindent\begin{minipage}[t]{.425\linewidth}
  \textbf{Example 2: $K \succeq 0, K = \zweightedsigmaki$}
  \begin{equation*}
    \begin{aligned}
      \underset{\zvec{\mu}, \zvec{\nu}, \zvec{\delta}, \lambda, t}{\text{min}} & \quad t                                                                                         \\
      \textit{subject to}                                                      & \quad trace(\zweightedsigmaki + \tau I_n) = c,                                                  \\
                                                                               & \quad \zweightedsigmaki \succeq 0                                                               \\
                                                                               & \quad \begin{pmatrix}
                                                                                         G(\zweightedsigmakitr) + \tau I_{n_{tr}}     & \zvec{e} + \zvec{\nu} + \lambda \zvec{y} \\
                                                                                         (\zvec{e} + \zvec{\nu} + \lambda \zvec{y})^T & t                                        \\
                                                                                       \end{pmatrix} \succeq 0                                                                   \\
                                                                               & \quad \zvec{\nu} \ge 0                                                                          \\
                                                                               & \quad \zvec{\delta} \ge 0                                                                       \\
    \end{aligned}
  \end{equation*}
\end{minipage}%
\begin{minipage}[t]{.6\linewidth}
  \textbf{Example 3: $K \succeq 0, K = \zweightedsigmaki, \zvec{\mu} \ge 0$}
  \begin{equation*}
    \begin{aligned}
      \underset{\zvec{\alpha}, t}{\text{max}} & \quad 2 \zvec{\alpha}^T \zvec{e} - c t                         \\
      \textit{subject to}                     & \quad t \ge 1/r_i \zvec{\alpha}^T G( K_{i, tr} ) \zvec{\alpha} \\
                                              & \quad t\ ge \zvec{\alpha}^T \zvec{\alpha}                      \\
                                              & \quad \zvec{\alpha}^T \zvec{y} = 0                             \\
                                              & \quad 0 \le \zvec{\alpha} \le C                                \\
                                              & \quad \forall ~ i \in [m]                                      \\
    \end{aligned}
  \end{equation*}
\end{minipage}

\noindent Notice how example 2 directly reduces to example 3 if $K_{single} \succeq 0$ which means automatically tuning the parameter $\tau = 1 / C$ for a 2-norm soft margin SVM with a singular kernel $K_{single}$. This suggests that even if we are not learning the kernel matrix (i.e., $K$ just equals $K_{single}$ and not $K = \zweightedsigmaki$), this approach can be used to tune the 2-norm soft margin parameter $\tau = 1 / C$ automatically, thus solving an important open problem of its time.

%% file: conclusion.tex
\section{Conclusion}

In this summary paper, we have described an application of semidefinite programming (SDP) in learning the kernel matrix used for support vector machines (SVM) algorithms. We first reviewed the basics of kernels-based learning algorithms, saw 3 different formulations of SVMs (i.e., hard margin, 1-norm soft margin, and 2-norm soft margin) and derived the a general dual form for all 3 cases which we called the generalized performance measure. Minimizing this generalized performance measure corresponded to maximizing the margin of the SVM classifier.  Next we studied the specialized yet commonplace form of SDPs and derived the corresponding dual formulation and saw that it is also an SDP. Using this property, we noted that the primal and dual can be jointly optimized using the same general purpose optimization solver, to more efficiently minimize the duality gap ($p^* - d^*$) used as a stopping criterion. This solver jointly optimized the primal-dual pair with complexity $O \big( (m + n_{tr})^2 n^{2.5} \big)$. At this point we had the tools to solve our main optimization problem, which was minimizing the generalized performance measure over a family of kernel matrices $\mathcal{K}$. Here we looked into 3 specific families, and demonstrated that when optimizing over a family of kernels that are strictly a positive linear combination of well-known kernel matrices, possibly from heterogeneous sources, leads to a significant improvement in both the complexity (which was $O \big( m n_{tr}^3 \big)$) and numerical stability of the optimal results. Finally, the formulations above were used to propose a method for finding the optimal 2-norm soft margin parameter $\tau = 1 / C$, which solved an open problem of its time.

%% file: appendix_a.tex
\section{Appendix A - Hard Margin Dual Formulation}
We start with:
\begin{equation*}
  \begin{aligned}
    \underset{\zvec{w}, b}{\text{min}} & \quad \langle \zvec{w}, \zvec{w} \rangle                                                         \\
    \textit{subject to}                & \quad y_i \big( \langle \zvec{w}, \Phi(\zvec{x_i}) \rangle + b \big) \ge 1 ~ \forall ~ i \in [n] \\
  \end{aligned}
\end{equation*}

\noindent and derive:
\begin{equation*}
  \begin{aligned}
    \underset{\zvec{\alpha}}{\text{max}} & \quad 2 \zvec{\alpha}^T \zvec{e} - \zvec{\alpha}^T \Big( G(K) +  \tau I_n \Big) \zvec{\alpha} \\
    \textit{subject to}                  & \quad \zvec{\alpha} \ge 0                                                                     \\
                                         & \quad \zvec{\alpha}^T \zvec{y} = 0                                                            \\
  \end{aligned}
\end{equation*}

\noindent where the constraints can be re-written in the proper form for deriving the dual formulation:
\begin{align*}
  1 - y_i \big( \langle \zvec{w}, \Phi(\zvec{x_i}) \rangle + b \big) \le 0 ~ \forall ~ i \in [n]
\end{align*}

Finally to simplify notation, we define $\zvec{z_i} = y_i \Phi(\zvec{x_i})$ and $Z = [\zvec{z_1}; \cdots; \zvec{z_n}]^T$ and write the $n$ constraints and write them together in the form
\begin{align*}
  \zvec{1} - Z \zvec{w} - b \zvec{y} \le 0
\end{align*}

\noindent Now we can define the Lagrangian as
\begin{align*}
  \mathcal{L}(Z, \zvec{\alpha}) = \underbrace{\langle \zvec{w}, \zvec{w} \rangle}_{\zvec{w}^T \zvec{w} = ||\zvec{w}||_2^2} + \zvec{\alpha}^T \big( \zvec{1} - Z \zvec{w} - b \zvec{y} \big)
\end{align*}

\noindent With the Lagrangian, our optimization problem becomes

\begin{equation*}
  \begin{aligned}
      \underset{\zvec{w}, b}{\text{min}}         \quad \underset{\zvec{\alpha} \ge 0}{\text{max}} \quad \mathcal{L}(Z, \zvec{\alpha}) \\
    = \underset{\zvec{\alpha} \ge 0}{\text{max}} \quad \underset{\zvec{w}, b}{\text{min}}         \quad \mathcal{L}(Z, \zvec{\alpha}) \\
  \end{aligned}
\end{equation*}

\noindent and this can be done because by Slater's condition we can see that the original minimization problem is strictly feasible. Alternative, we can see that the objective function is convex in $\zvec{w}$ and concave in $\zvec{\alpha}$ and so the switching of $min$ and $max$ is justified. After swapping $max$ and $min$, we can solve the inner optimization problem by taking derivative and setting to $0$:
\[
  \begin{cases}
    \frac{\partial \mathcal{L}(Z, \zvec{\alpha})}{\partial \zvec{w}} &= 2 \zvec{w} - \frac{\partial ( \zvec{\alpha}^T Z \zvec{w} )}{\partial \zvec{w}} = 2 \zvec{w} - \frac{\partial ( \zvec{w}^T Z^T \zvec{\alpha} )}{\partial \zvec{w}} = 2 \zvec{w} - Z^T \zvec{\alpha} = 0 \\
    \frac{\partial \mathcal{L}(Z, \zvec{\alpha})}{\partial b}        &= - \zvec{\alpha}^T \zvec{y} = 0                                                                                                                                                                         \\
  \end{cases}
  \implies
  \begin{cases}
    \zvec{w^*}               &= \frac{Z^T \zvec{\alpha}}{2} \\
    \zvec{\alpha}^T \zvec{y} &= 0                           \\
  \end{cases}
\]

Plugging back into the inner optimization function (the $min$ problem), we get
\begin{align*}
                                                        \mathcal{L}(Z, \zvec{\alpha})               &= \langle \zvec{w}, \zvec{w} \rangle + \zvec{\alpha}^T \big( \zvec{1} - Z \zvec{w} - b \zvec{y} \big)                       \\
  \overset{\zvec{w^*} = Z^T \zvec{\alpha} / 2}{\implies} \mathcal{L}_{\zvec{w^*}}(Z, \zvec{\alpha}) &= \frac{\zvec{\alpha}^T Z Z^T \zvec{\alpha}}{4} + \zvec{\alpha}^T \big( \zvec{1} - Z (\frac{Z^T \zvec{\alpha}}{2})^T \big)    \\
                                                                                                    &= \zvec{\alpha}^T \zvec{1} - \frac{\zvec{\alpha}^T Z Z^T \zvec{\alpha}}{2}                                                    \\
\end{align*}

\noindent note that we still have the conditions $\zvec{\alpha} \ge 0$ and $\zvec{\alpha}^T \zvec{y} = 0$ that are conditions for the outer optimization function (the $max$ problem). So now the overall optimization problem becomes:

\begin{equation*}
  \begin{aligned}
    \underset{\zvec{\alpha} \ge 0}{\text{max}} & \quad \zvec{\alpha}^T \zvec{1} - \frac{\zvec{\alpha}^T Z Z^T \zvec{\alpha}}{2} \\
    \textit{subject to}                        & \quad \zvec{\alpha} \ge 0                                                      \\
                                               & \quad \zvec{\alpha}^T \zvec{y} = 0                                             \\
  \end{aligned}
\end{equation*}

\noindent note that we can re-write $\zvec{\alpha}^T \zvec{1} - \frac{\zvec{\alpha}^T Z Z^T \zvec{\alpha}}{2}$ as $1 \zvec{\alpha}^T \zvec{1} - \zvec{\alpha}^T Z Z^T \zvec{\alpha}$ without affect the maximizer of the optimization problem. Furthermore, note that $Z Z^T = y_i y_j k(\zvec{x_i}, \zvec{x_j})$ which is what we defined to be $G(K)$. So the final formulation becomes:

\begin{equation*}
  \begin{aligned}
    \underset{\zvec{\alpha}}{\text{max}} & \quad 2 \zvec{\alpha}^T \zvec{e} - \zvec{\alpha}^T \Big( G(K) +  \tau I_n \Big) \zvec{\alpha} \\
    \textit{subject to}                  & \quad \zvec{\alpha} \ge 0                                                                     \\
                                         & \quad \zvec{\alpha}^T \zvec{y} = 0                                                            \\
  \end{aligned}
\end{equation*}

\noindent which is what we set out to get as the derived dual formulation. \qed

The same process above can extended to derive the dual derivation of the 1-norm and 2-norm soft margins with slack variables.

%% file: appendix_b.tex
\newcommand{\zomegaktr}{\omega_{C,\tau}(K_{tr})}
\newcommand{\zlagrangian}{\mathcal{L}(\zvec{\alpha}, \zvec{\nu}, \zvec{\delta}, \lambda)}

\section{Appendix B - SDP Formulation for General Optimization Problem Assuming $K \succeq 0, K = \sum_{i = 1}^m \mu_i K_{i, tr}$}

\noindent We begin by substituting $\zomegaktr$ into the general optimization problem to get

\begin{equation*}
  \begin{aligned}
    \underset{K}{\text{min}} & \quad \underset{\zvec{\alpha}}{\text{max}} \quad 2 \zvec{\alpha}^T \zvec{e} - \zvec{\alpha}^T \Big( G(K_{tr}) +  \tau I_{n_{tr}} \Big) \zvec{\alpha} \\
    \textit{subject to}      & \quad 0 \le \zvec{\alpha} \le C                                                                                                                      \\
                             & \quad \zvec{\alpha}^T \zvec{y} = 0                                                                                                                   \\
                             & \quad trace(K) = c                                                                                                                                   \\
  \end{aligned}
\end{equation*}

\zbvspace

Assuming $K_{tr} \succeq 0$ we have $G(K_{tr}) \succeq 0$, and since $\tau \ge 0$ we have $\tau I_{n_{tr}} \succeq 0$ and so $G(K_{tr}) + \tau I_{n_{tr}} \succeq 0$ (can be generalized to the semidefinite case). We know from Section \ref{section_algorithms_for_learning_kernels} that $\zomegaktr$ is convex in $K_{tr}$ and thus in $K$, and given the remainder of the convex constraints, the optimization problem is certainly convex in $K$ and can be rewritten as

\begin{equation*}
  \begin{aligned}
    \underset{K}{\text{min}} & \quad t                                                                                                                                                    \\
    \textit{subject to}      & \quad t \ge \underset{\zvec{\alpha}}{\text{max}} \quad 2 \zvec{\alpha}^T \zvec{e} - \zvec{\alpha}^T \Big( G(K_{tr}) +  \tau I_{n_{tr}} \Big) \zvec{\alpha} \\
                             & \quad 0 \le \zvec{\alpha} \le C                                                                                                                            \\
                             & \quad \zvec{\alpha}^T \zvec{y} = 0                                                                                                                         \\
                             & \quad trace(K) = c                                                                                                                                         \\
  \end{aligned}
\end{equation*}

\zbvspace

\noindent We now express the first constraint using its dual formulation by defining the Lagrangian of the maximization problem as

\begin{align*}
  \zlagrangian = 2 \zvec{\alpha}^T \zvec{e} - \zvec{\alpha}^T \Big( G(K_{tr}) +  \tau I_{n_{tr}} \Big) \zvec{\alpha} + 2 \zvec{\nu}^T \zvec{\alpha} + 2 \zvec{\delta}^T (C \zvec{e} - \zvec{\alpha}) + \underbrace{2 \lambda \zvec{\alpha}^T \zvec{y}}_{= 2 \lambda \zvec{y}^T \zvec{\alpha}}
\end{align*}

% \zbvspace

\noindent where $\lambda \in \mathbb{R}$ and $\zvec{\nu}, \zvec{\delta} \in \mathbb{R}^{n_{tr}}$. By duality we have

\begin{align*}
  \zomegaktr = \underset{\zvec{\alpha}}{\text{max}} ~ \underset{\zvec{\nu} \ge 0, \zvec{\delta} \ge 0, \lambda}{\text{min}} ~ \zlagrangian = \underset{\zvec{\nu} \ge 0, \zvec{\delta} \ge 0, \lambda}{\text{min}} ~ \underset{\zvec{\alpha}}{\text{max}} ~ \zlagrangian
\end{align*}

\noindent Similar to Appendix A, we compute $\frac{\partial \zlagrangian}{\partial \zvec{\alpha}}$ and considering $G(K_{tr}) + \tau I_{n_{tr}} \succeq 0$ we get

\begin{align*}
           \frac{\partial \zlagrangian}{\partial \zvec{\alpha}} &= 2 \zvec{e} + 2 \big( G(K_{tr}) + \tau I_{n_{tr}} \big) \zvec{\alpha} + 2 \zvec{\nu} - 2 \zvec{\delta} + 2 \lambda \zvec{y} = 0 \\
  \implies \zvec{\alpha}                                        &= \big( G(K_{tr}) + \tau I_{n_{tr}} \big)^{-1} (\zvec{e} + \zvec{\nu} - \zvec{\delta} + \lambda \zvec{y})                        \\
\end{align*}

\noindent and so the dual problem of the maximization problem becomes

\begin{equation*}
  \begin{aligned}
    \zomegaktr = \underset{\zvec{\nu} \ge 0, \zvec{\delta} \ge 0, \lambda}{\text{min}} & \quad (\zvec{e} + \zvec{\nu} - \zvec{\delta} + \lambda \zvec{y})^T \big( G(K_{tr}) + \tau I_{n_{tr}} \big)^{-1} (\zvec{e} + \zvec{\nu} - \zvec{\delta} + \lambda \zvec{y}) + 2 C \zvec{\delta}^T \zvec{e} \\
    \textit{subject to}                                                                & \quad \zvec{\nu} \ge 0                                                                                                                                                                                    \\
                                                                                       & \quad \zvec{\delta} \ge 0                                                                                                                                                                                 \\
  \end{aligned}
\end{equation*}

\noindent This implies that for any $t > 0$ the constraint $\zomegaktr$ holds $\iff ~ \exists \zvec{\nu} \ge 0 ~ \& ~ \zvec{\delta} \ge 0$ such that

\begin{align*}
  (\zvec{e} + \zvec{\nu} - \zvec{\delta} + \lambda \zvec{y})^T \big( G(K_{tr}) + \tau I_{n_{tr}} \big)^{-1} (\zvec{e} + \zvec{\nu} - \zvec{\delta} + \lambda \zvec{y}) + 2 C \zvec{\delta}^T \zvec{e} \le t
\end{align*}

\noindent or equivalently (using the Schur complement lemma)

\begin{align*}
  \begin{pmatrix}
    G(K_{tr}) + \tau I_{n_{tr}}                               & \zvec{e} + \zvec{\nu} - \zvec{\delta} + \lambda \zvec{y} \\
    (\zvec{e} + \zvec{\nu} - \zvec{\delta} + \lambda \zvec{y})^T & t - 2C \zvec{\delta}^T \zvec{e}                       \\
  \end{pmatrix} \succeq 0                                                                                                \\
\end{align*}

\noindent Plugging back into the form we started with yields

\begin{equation*}
  \begin{aligned}
    \underset{K, \zvec{\nu}, \zvec{\delta}, \lambda, t}{\text{min}} & \quad t                                                                                                                   \\
    \textit{subject to}                                          & \quad trace(K) = c,                                                                                                          \\
                                                                 & \quad K \in \mathcal{K}                                                                                                      \\
                                                                 & \quad \begin{pmatrix}
                                                                           G(K_{tr}) + \tau I_{n_{tr}}                               & \zvec{e} + \zvec{\nu} - \zvec{\delta} + \lambda \zvec{y} \\
                                                                           (\zvec{e} + \zvec{\nu} - \zvec{\delta} + \lambda \zvec{y})^T & t - 2C \zvec{\delta}^T \zvec{e}                       \\
                                                                         \end{pmatrix} \succeq 0                                                                                                \\
                                                                 & \quad \zvec{\nu} \ge 0                                                                                                       \\
                                                                 & \quad \zvec{\delta} \ge 0                                                                                                    \\
  \end{aligned}
\end{equation*}

\noindent which is what we set out to get. As mentioned earlier, the objective function as well as all constraints above are convex (notice again that $\zvec{\nu} \ge 0 \iff diag(\zvec{\nu}) \succeq 0$ which means that contraint is an LMI and a proper SDP constraint; likewise for $\zvec{\nu} \ge 0$.), and so the general optimization problem is now written in SDP form. \qed

%% file: appendix_c.tex
\newcommand{\maxconstraint}{\zvec{\alpha}: 0 \le \zvec{\alpha} \le C, \zvec{\alpha}^T y = 0}
\newcommand{\minconstraint}{\zvec{\mu}: \zvec{\mu} \ge 0, \zvec{\mu}^T \zvec{r} = c}

\section{Appendix B - SDP Formulation for General Optimization Problem Assuming $K \succeq 0, K = \sum_{i = 1}^m \mu_i K_{i, tr}, \zvec{\mu} \ge 0$}

\noindent We begin by substituting $\zomegaktr$ into the general optimization problem to get

\begin{equation*}
  \begin{aligned}
    \underset{K}{\text{min}} & \quad \underset{\maxconstraint}{\text{max}} \quad 2 \zvec{\alpha}^T \zvec{e} - \zvec{\alpha}^T \Big( G(K_{tr}) + \tau I_{n_{tr}} \Big) \zvec{\alpha} \\
    \textit{subject to}      & \quad trace(K) = c                                                                                                                                    \\
                             & \quad K \succeq 0                                                                                                                                     \\
                             & \quad K = \sum_{i = 1}^m \mu_i K_{i, tr}                                                                                                              \\
                             & \quad \zvec{\mu} \ge 0                                                                                                                                \\
  \end{aligned}
\end{equation*}

% \zbvspace

\noindent where the second constraint $K \succeq 0$ can be omitted as it is implied by the last two constraints. Recall that $\zvec{r} \ in \mathbb{R}^m$ and $trace(K_i) = r_i$. So the above can be re-written as:

\begin{equation*}
  \begin{aligned}
    \underset{\mu}{\text{min}} & \quad \underset{\maxconstraint}{\text{max}} \quad 2 \zvec{\alpha}^T \zvec{e} - \zvec{\alpha}^T \Big( G(\sum_{i = 1}^m \mu_i K_{i, tr}) + \tau I_{n_{tr}} \Big) \zvec{\alpha} \\
    \textit{subject to}        & \quad\zvec{\mu}^t \zvec{r} = c                                                                                                                                                \\
                               & \quad\zvec{\mu}^t \ge 0                                                                                                                                                       \\
  \end{aligned}
\end{equation*}

\noindent where $K_{i, tr} = K_i(1 : n_{tr}, 1 : n_{tr})$. This can be further reduced as

\begin{equation*}
  \begin{aligned}
                  & \underset{\minconstraint}{\text{min}} \quad \underset{\maxconstraint}{\text{max}} \quad 2 \zvec{\alpha}^T \zvec{e} - \zvec{\alpha}^T \Big( G                     ( \sum_{i = 1}^m \mu_i K_{i, tr} )                       + \tau I_{n_{tr}} \Big) \zvec{\alpha}    \\
                = & \underset{\minconstraint}{\text{min}} \quad \underset{\maxconstraint}{\text{max}} \quad 2 \zvec{\alpha}^T \zvec{e} - \zvec{\alpha}^T \Big( \text{diag}(\zvec{y}) ( \sum_{i = 1}^m \mu_i K_{i, tr} ) \text{diag}(\zvec{y}) + \tau I_{n_{tr}} \Big) \zvec{\alpha}    \\
                = & \underset{\minconstraint}{\text{min}} \quad \underset{\maxconstraint}{\text{max}} \quad 2 \zvec{\alpha}^T \zvec{e} - \sum_{i = 1}^m \mu_i \zvec{\alpha}^T \text{diag}(\zvec{y}) K_{i, tr} \text{diag}(\zvec{y}) \zvec{\alpha} - \tau \zvec{\alpha}^T \zvec{\alpha} \\
    \overset{r1}= & \underset{\maxconstraint}{\text{max}} \quad \underset{\minconstraint}{\text{min}} \quad 2 \zvec{\alpha}^T \zvec{e} - \sum_{i = 1}^m \mu_i \zvec{\alpha}^T \text{diag}(\zvec{y}) K_{i, tr} \text{diag}(\zvec{y}) \zvec{\alpha} - \tau \zvec{\alpha}^T \zvec{\alpha} \\
    \overset{r2}= & \underset{\maxconstraint}{\text{max}} \quad \underset{\minconstraint}{\text{min}} \quad 2 \zvec{\alpha}^T \zvec{e} - \sum_{i = 1}^m \mu_i \zvec{\alpha}^T                    G( K_{i, tr} )                     \zvec{\alpha} - \tau \zvec{\alpha}^T \zvec{\alpha} \\
                = & \underset{\maxconstraint}{\text{max}} \quad \bigg[ 2 \zvec{\alpha}^T \zvec{e} - \tau \zvec{\alpha}^T \zvec{\alpha} - \underset{\minconstraint}{\text{max}} \bigg ( \sum_{i = 1}^m \mu_i \zvec{\alpha}^T G( K_{i, tr} ) \zvec{\alpha} \bigg) \bigg]                 \\
                = & \underset{\maxconstraint}{\text{max}} \quad \bigg[ 2 \zvec{\alpha}^T \zvec{e} - \tau \zvec{\alpha}^T \zvec{\alpha} - \underset{i}{\text{max}}              \bigg ( \frac{c}{r_i}        \zvec{\alpha}^T G( K_{i, tr} ) \zvec{\alpha} \bigg) \bigg]                 \\
  \end{aligned}
\end{equation*}

\noindent with reasoning:
\begin{itemize}
  \item r1: interchanging the $min$ \& $max$ is justified because the objective is linear in $\zvec{\mu}$ (hence convex) and concave in $\zvec{\alpha}$
  \item r2: $G( K_{i, tr} ) = \text{diag}(\zvec{y}) K_{i, tr} \text{diag}(\zvec{y})$ from definitin where $G(K)$ is defined by $G_{ij}(K) = y_i y_j [K]_{ij} = y_i y_j k(\zvec{x_i}, \zvec{x_j})$
\end{itemize}

This can finally be fomulated as:

\begin{equation*}
  \begin{aligned}
    \underset{\zvec{\alpha}, t}{\text{max}} & \quad 2 \zvec{\alpha}^T \zvec{e} - \tau \zvec{\alpha}^T \zvec{\alpha} - c t \\
    \textit{subject to}                     & \quad t \ge 1/r_i \zvec{\alpha}^T G( K_{i, tr} ) \zvec{\alpha}              \\
                                            & \quad \zvec{\alpha}^T \zvec{y} = 0                                          \\
                                            & \quad 0 \le \zvec{\alpha} \le C                                             \\
                                            & \quad \forall ~ i \in [m]                                                   \\
  \end{aligned}
\end{equation*}

\qed